\documentclass[a4paper,twoside]{article}

\usepackage{calc}
\usepackage{amssymb}
\usepackage{amstext}
\usepackage{amsmath}
\usepackage{amsthm}
\usepackage{multicol}
\usepackage{pslatex}
\usepackage{apalike}

\usepackage{amssymb}
\usepackage{graphicx}

\newcommand{\keywords}[1]{\par\addvspace\baselineskip
\noindent\keywordname\enspace\ignorespaces#1}

\usepackage{times}
\usepackage{algorithm}
\usepackage{algorithmic}

\usepackage{float}
\usepackage[english]{babel}

\usepackage[caption=false,font=footnotesize]{subfig}

\usepackage{enumitem}
\usepackage{hyperref}

\newenvironment{lyxlist}[1]
{\begin{list}{}
		{\settowidth{\labelwidth}{#1}
			\setlength{\leftmargin}{\labelwidth}
			\addtolength{\leftmargin}{\labelsep}
			}}
	{\end{list}}

\usepackage{SCITEPRESS}     

\begin{document}

\title{Optimized Linear Imputation}

\author{\authorname{Yehezkel S. Resheff\sup{1,2}, Daphna Weinshall\sup{1}}
\affiliation{\sup{1}School of Computer Science and Engineering, The Hebrew University of Jerusalem}
\affiliation{\sup{2}Edmond and Lily Safra Center for Brain Sciences, The Hebrew University of Jerusalem}
\email{\{heziresheff, daphna\}@cs.huji.ac.il}
}

\keywords{Imputation}

\abstract{Often in real-world datasets, especially in high dimensional data, some feature values are missing. Since most data analysis and statistical methods do not handle gracefully missing values, the first step in the analysis requires the imputation of missing values. Indeed, there has been a long standing interest in methods for the imputation of missing values as a pre-processing step. One recent and effective approach, the IRMI stepwise regression imputation method, uses a linear regression model for each real-valued feature on the basis of all other features in the dataset. However, the proposed iterative formulation lacks convergence guarantee. Here we propose a closely related method, stated as a single optimization problem and a block coordinate-descent solution which is guaranteed to converge to a local minimum. Experiments show results on both synthetic and benchmark datasets, which are comparable to the results of the IRMI method whenever it converges. However, while in the set of experiments described here IRMI often diverges, the performance of our methods is shown to be markedly superior in comparison with other methods.}

\onecolumn \maketitle \normalsize \vfill

\section{\uppercase{Introduction}}
\label{sec:introduction}

Missing data imputation is an important part of data preprocessing and cleansing
\cite{horton2007much,pigott2001review}, since the vast majority of commonly applied supervised
machine learning and statistical methods for classification rely on complete data
\cite{garcia2010pattern}.  The most common option for many applications is to discard complete
records in which there are any missing values. This approach is insufficient for several reasons:
first, when missing values are not missing at random
\cite{little1988test,heitjan1996distinguishing}, discarding these records may bias the resulting
analysis \cite{little2014statistical}.  Other limitations include the loss of information when
discarding the entire record. Furthermore, when dealing with datasets with either a small number of
records or a large number of features, omitting complete records when any feature value is missing
may result in insufficient data for the required analysis.

Early methods for data imputation include methods for replacing a missing value by the mean or
median of the feature value across records \cite{engels2003imputation,donders2006review}.  While
these values may indeed provide a ``good guess'' when there is no information present, this is often
not case.  Namely, for each missing feature value there are other non-missing values in the same
record. It is likely therefore (or indeed, we assume) that other features contain information
regarding the missing feature, and imputation should therefore take into account known feature
values in the same record. This is done by subsequent methods.

Multiple imputation (see \cite{rubin1996multiple} for a detailed review) imputes several sets of
missing values, drawn from the posterior distribution of the missing values under a given model,
given the data. Subsequent processing is then to be performed on each version of the imputed data,
and the resulting multiple sets of model parameters are combined to produce a single result.  While
extremely useful in traditional statistical analysis and public survey data, it may not be feasible
in a machine learning setting. First, the run-time cost of performing the analysis on several copies
of the full-data may be prohibitive. Second, being a model-based approach it depends heavily on the
type and nature of the data, and can't be used as an out-of-the-box pre-processing step. More
importantly though, while traditional model parameters may be combined between versions of the
imputed data (regression coefficients for instance), many modern machine learning methods do not
produce a representation that is straightforward to combine (consider the parameters of an
Artificial Neural Network or a Random Forest for example
\footnote{In this case it would be perhaps more natural to train the model using data pooled over
	the various copies of the completed data rather than train separate models and average the
	resulting parameters and structure.  This is indeed done artificially in methods such as
	denoinsing neural nets \cite{vincent2010stacked}, and has been known to be useful for data
	imputation \cite{duan2014deep}.  }).

In \cite{raghunathan2001multivariate}, a method for imputation on the basis of a sequence of
regression models is introduced. This method, popularized under the acronym MICE
\cite{buuren2011mice,van1999flexible}, uses a non-empty set of complete features which are known in
all the records as its base, and iteratively imputes one feature at a time on the basis of the
completed features up to that point. Since each step produces a single complete feature, the number
of iterations needed is exactly the number of features that have a missing value in at least one
record.  The drawbacks of this method are twofold. First, there must be at least one complete
feature to be used as the base. More importantly though, the values imputed at the $i-th$ step can
only use a regression model that includes the features which were originally full or those imputed
in the $i-1$ first steps. Ideally, the regression model for each feature should be able to use all
other feature values.

The IRMI method \cite{templ2011iterative} goes one step further by building a sequence of
regression models for each feature that can use all other feature values as needed. This iterative
method initially uses a simple imputation method such as median imputation. In each iteration it
computes for each feature the linear regression model based on all other feature values, and then
re-imputes the missing values based on these regression models. The process is terminated upon
convergence or after a per-determined number of iterations (Algorithm~\ref{alg:1}). The authors
state that although they do not have a proof of convergence, experiments show fast convergence in
most cases.

In Section~\ref{sec:oli} we present a novel method of Optimized Linear Imputation (OLI). The OLI
method is related in spirit to IRMI in that it performs a linear regression imputation for the
missing values of each feature, on the basis of all other features. Our method is defined by a
single optimization objective which we then solve using a block coordinate-descent method. Thus our
method is guaranteed to converge, which is its most important advantage over IRMI. We further show
that our algorithm may be easily extended to use any form of regularized linear regression.

In Section~\ref{sec:simul} we ompare the OLI method to the IRMI, MICE and Median Imputation (MI)
methods.  Using the same simulation studies as in the original IRMI paper, we show that the results
of OLI are rather similar to the results of IRMI.  With real datasets we show that our method
usually outperforms the alternatives MI and MICE in accuracy, while providing comparable results to
IRMI. However, IRMI did not converge in many of these experimentsm while our method always provided
good results.

\begin{algorithm*}[tbh]
	input: 
	\begin{itemize}
		\setlength\itemsep{-0.5em}
		\item $X$ - data matrix of size $N\times (d+1)$ containing $N$ samples and
		$d$ features
		\item $m$ - missing data mask 
		\item $max\_iter$ - maximal number of iterations
	\end{itemize}
	output:
	\begin{itemize}
		\item Imputation values \\
		
	\end{itemize}
	
	\begin{algorithmic}[1] 
		
		\STATE{$\tilde X := median\_impute(X)$} 
		\COMMENT{assigns each missing value the median of its column}
		\WHILE{not converged and under $max\_iter$ iterations} 
		\FOR{i := 1...d} 
		\STATE{regression = linear\_regression($\tilde X_{-i}[!m_i], \tilde X_{i}[!m_i]$)}
		
		\STATE {$\tilde X_{i}[m_i]$ = regression.predict($\tilde X_{-i}[m_i]$) }
		
		\ENDFOR		
		\ENDWHILE
		\STATE {\textbf{return} {$\tilde X-X$}}
	\end{algorithmic}
	
	\protect\caption{the IRMI method for imputation of real-valued features (see \cite{templ2011iterative}
		for more details)}
	\label{alg:1}
\end{algorithm*}

\section{uppercase{OLI method}}
\label{sec:oli}

\subsection {Notation} 

We start by listing the notation used throughout the paper. 
\begin{lyxlist}{00.00.0000}
	\item [{$N$}] Number of samples 
	\item [{$d$}] Number of features
	\item [{$x_{i,j}$}] The value of the $j-th$ feature in the $i-th$ sample
	\item [{$m_{i,j}$}] Missing value indicators:
	\begin{lyxlist}{00.00.0000}
		\item [{$m_{i,j}=\begin{cases}
			1 & x_{i,j}\ is\ missing\\
			0 & otherwise
			\end{cases}$}]~
	\end{lyxlist}
	\item [{$m_i$}] Indicator vector of missing values for for the $i-th$ feature
\end{lyxlist}

The following notation is used in the algorithms' pseudo-code:

\begin{lyxlist}{00.00.0000}
	\item [{$A[m]$}] The rows of a matrix (or column vector) $A$ where the boolean
	mask vector $m$ is $True$
	\item [{$A[!m]$}] The rows of a matrix (or column vector) $A$ where the boolean
	mask vector $m$ is $False$
	\item [linear\_regression($X$, $y$)] A linear regression from the columns of the matrix
	$X$ to the target vector $y$, having the following fields:
	\begin{lyxlist}{00.00.0000}
		\item [.parameters:] parameters of the fitted model.
		\item [.predict($X$):] the target column $y$ as predicted by the fitted model. 
	\end{lyxlist}
	
\end{lyxlist}
\subsection {Optimization problem}

We formulate the linear imputation as a single optimization problem.
First we construct a design matrix:

\begin{equation}
X=\begin{bmatrix} &  &  & 1\\
& [x_{i,j}(1-m_{i,j})] &  & \vdots\\
&  &  & 1
\end{bmatrix}
\end{equation}

\noindent where the constant-$1$ rightmost column is used for the intercept terms in the subsequent
regression models. Multiplying the data values $x_{i,j}$ by $(1-m_{i,j})$ simply sets all missing
values to zero, keeping non-missing values as they are.

Our approach aims to find consistent missing value imputations and regression coefficients as a
single optimization problem. By consistent we mean that (a) the imputations are the values obtained
by the regression formulas, and (b) the regression coefficients are the values that would be computed
after the imputations. We propose the following optimization formulation:

\begin{equation}
\begin{cases}
\underset{A,M}{\min} & ||(X+M)A-(X+M)||_{F}^{2}\\
s.t. & m_{i,j}=0\Rightarrow M_{i,j}=0\\
& M_{i,d+1}=0\ \forall i\\
& A_{i,i}=0\ \ \ \ \ \ i=1...d\\
& A_{i,d+1} = \delta_{i, d+1}\ \ \ \forall i 
\end{cases}
\label{eq:optimization_problem}
\end{equation}

\noindent where $||\ldotp||_{F}$ is the Frobenius norm. 

Intuitively, the objective that we minimize measures the square error of reconstruction of the
imputed data $(X+M)$, where each feature (column) is approximated by a linear combination of all
other features plus a constant (that is, linear regression of the remaining imputed data). The
imputation process by which $M$ is defined is guaranteed to leave the non-missing values in $X$
intact, by the first and second constraints which make sure that only missing entries in $X$ have a
corresponding non-zero value in $M$. Therefore:

\[
(X+M)=\begin{cases}
M & for\ missing\ values\\
X & for\ non\ missing\ values
\end{cases}
\]

The regression for each feature is further constrained to use only \textbf{other }features, by
setting the diagonal values of $A$ to zero (the third constraint). The forth constraint makes sure
that the constant-$1$ rightmost column of the design matrix is copied as-is and therefore does not
impact the objective. 

We note that all the constraints set variables to constant values, and therefore this can be seen as
an unconstrained optimization problem on the remaining set of variables. This set includes the
non-diagonal elements of $A$ and the elements of $M$ corresponding to missing values in $X$.  We
further note that this is not a convex problem in $A,M$ since it contains the $MA$ factor. In the
next section we show a solution to this problem that is guaranteed to converge to a local minimum.

\subsection {Block coordinate descent solution}

We now develop a coordinate descent solution for the proposed optimization problem. Coordinate descent (and more specifically alternating least squares; see for example \cite{hope2016ballpark}) algorithms are extremely common in machine learning and statistics, and while don't guarantee convergence to a global optimum, they often preform well in practise.
 
As stated above, our problem is an unconstrained optimization problem over the following set of variables:

\[
\{A_{i,j}|i,j=1,..,d;i\neq j\}\cup\{M_{i,j}|m_{i,j}=1\}
\]

\begin{algorithm*}[tbh]
	input: 
	\begin{itemize}
		\setlength\itemsep{-0.5em}
		\item $X_0$ - data matrix of size $N\times d$ containing $N$ samples and
		$d$ features
		\item $m$- missing data mask 
	\end{itemize}
	output:
	\begin{itemize}
		\item Imputation values\\
		
	\end{itemize}
	\begin{algorithmic}[1] 
		
		\STATE{$X := median\_impute(X_0)$}
		
		\STATE{$M := zeros(N, d)$}
		\STATE{$A := zeros(d, d)$}  
		\WHILE {not converged} 
		\FOR{$i := 1...d$} 
		\STATE {$\beta := linear\_regression(X_{-i}, X_{i}).parameters$}				
		\STATE {$A_{i}:=[\beta_{1},...,\beta_{i-1},0,\beta_{i},...,\beta_{d}]^{T}$}		
		\ENDFOR`	
		\WHILE {not converged} 
		\STATE {$M := M- \alpha[(X+M)A-(X+M)](A-I)^{T}$} 
		\STATE {$M[!m] := 0$} 
		\ENDWHILE
		\STATE {$X := X + M$}	
		\ENDWHILE
		\STATE {\textbf{return} {$M$}}
	\end{algorithmic}
	
	\protect\caption{Optimized Linear Imputation (OLI)}
	\label{alg:2}
\end{algorithm*}

\noindent Keeping this in mind, we use the following objective function:

\begin{align}
	L(A,M)&=||(X+M)A-(X+M)||_{F}^{2}\label{eq:objective} \\
	&=\sum_{i=1}^{d}||(X+M)_{-i}\beta_{i}-(X+M)_{i}||_{F}^{2}\label{eq:i_itemized_objective}
\end{align}

\noindent where $C_{-i}$ denotes the matrix $C$ without its $i-th$ column, $C_{i}$ the $i-th$
column, and $\beta_{i}$ the $i-th$ column of $A$ without the $i-th$ element (recall that the $i-th$
element of the $i-th$ column of $A$ is always zero). The term $(X+M)_{-i}\beta_{i}$ is therefore a
linear combination of all but the $i-th$ column of the matrix $(X+M)$. The sum in (\ref{eq:i_itemized_objective}) is over the first $d$ columns only, since the term added by the rightmost column is zero (see fourth constraint in (\ref{eq:optimization_problem})).

We now suggest the following coordinate descent algorithm for the minimization of the objective
(\ref{eq:objective}) (the method is summarized in Algorithm~\ref{alg:2}):
\begin{enumerate}
	\item 
	Fill in missing values using median/mean (or any other) imputation 
	\item Repeat until convergence:
	
	\begin{enumerate}
		\item Minimize the objective (\ref{eq:objective}) w.r.t. A (compute the columns of the matrix $A$)
		\item Minimize the objective (\ref{eq:objective}) w.r.t. M (compute the missing values entries in matrix M)
	\end{enumerate}
	\item Return $M$ \footnote{Alternatively, in order to stay close in spirit to the linear IRMI method, we may prefer to use $(X+M)A$ as the imputed data. }
	
\end{enumerate}
As we will show shortly, step (a) in the iterative part of the proposed algorithm reduces to
calculating the linear regression for each feature on the basis of all other features, essentially
the same as the first step in the IRMI algorithm \cite{templ2011iterative} Algorithm~\ref{alg:1}.
Step (b) can be solved either as a system of linear equations or in itself as an iterative
procedure, by gradient descent on (\ref{eq:objective}) w.r.t $M$ using (\ref{eq:dLdM}).

First, we show that step (a) reduces to linear regression. Taking the derivatives of
(\ref{eq:i_itemized_objective}) w.r.t the non-diagonal elements of column $i$ of $A$ we have:

\begin{equation*}
	\frac{\partial L}{\partial \beta_{i}}=2(X+M)_{-i}^{T}[(X+M)_{-i}\beta_{i}-(X+M)_{i}]
\end{equation*}

\noindent Setting the partial derivatives to zero gives:

\begin{align*}
	&(X+M)_{-i}^{T}[(X+M)_{-i}\beta_{i}-(X+M)_{i}]{=}0 \cr
	\Rightarrow & \beta_{i}=((X+M)_{-i}^{T}(X+M)_{-i})^{-1}(X+M)_{-i}^{T}(X+M)_{i}
\end{align*}

\noindent which is exactly the linear regression coefficients for the $i-th$ feature from all
other (imputed) features, as claimed.

\noindent Next, we obtain the derivatives of the objective function w.r.t $M$:

\begin{equation}
\nabla_{M} = \frac{\partial L}{\partial M}=2[(X+M)A-(X+M)](A-I)^{T}
\label{eq:dLdM}
\end{equation}

\noindent leading to the following gradient descent algorithm for step (b):
\noindent step (b), Repeat until convergence:

\begin{enumerate}[label=(\roman*)]
	\item $M:=M-\alpha\nabla_{M}L(A,M)$
	\item $\forall_{i,j}:\ M_{i,j}=M_{i,j}m_{i,j}$
\end{enumerate}
where $\alpha$ is a predefined step size and the gradient is given by (\ref{eq:dLdM}). Step (ii)
makes sure that only missing values are assigned imputation values\footnote{Note that this is not a
	projection step.  Recall that the optimization problem is only over elements $M_{ij}$ where
	$x_{ij}$ is a missing value, encoded by $m_{ij}=1$.  The element-wise multiplication of $M$ by $m$
	guarantees that all other elements of $M$ are assigned $0$. Effectively, the gradient descent
	procedure does not treat them as independent variables, as required.}.

Our proposed algorithm uses a gradient descent procedure for the minimization of the objective
(\ref{eq:objective}) w.r.t $M$. Alternatively, one could use a closed form solution by directly
setting the partial derivative to zero. More specifically, let

\begin{equation}
\frac{\partial L}{\partial M}{=}0\label{eq:dLdM--0}
\end{equation}

\noindent Substituting (\ref{eq:dLdM}) into (\ref{eq:dLdM--0}), we get

\[
M(A-I)(A-I)^{T}=-X(A-I)(A-I)^{T}
\]

\noindent which we rewrite as:

\begin{equation}
MP=Q\label{eq:MP=00003DQ}
\end{equation}

\noindent with the appropriate matrices $P,Q$. Now, since only elements of $M$ corresponding to
missing values of $X$ are optimization variables, only these elements must be set to zero in the
derivative (\ref{eq:dLdM--0}), and hence only these elements must obey the equality
(\ref{eq:MP=00003DQ}).  Thus, we have:

\[
(MP)_{i,j}=Q_{i,j}\ \ \forall i,j|m_{i,j}=1
\]

\noindent which is a system of $\sum\limits_{i,j} m_{i,j}$ linear equations in $\sum\limits_{i,j}
m_{i,j}$ variables.

\subsection {Discussion}

In order to better understand the difference between the IRMI and OLI methods, we rewrite the IRMI
iterative method \cite{templ2011iterative} using the same notation as used for our method. We start
by defining an error matrix:

\[
E=(X+M)A-(X+M)
\]

\noindent $E$ is the error matrix of the linear regression models on the basis of the imputed
data. Unlike our method, however, IRMI considers the error only in the non-missing values of the
data, leading to the following objective function:

\[
L(M,A)=\underset{i,j|m_{i,j}=0}{\sum}E_{i,j}^{2}
\]

In order to minimize this loss function, at each step the IRMI method (Algorithm~\ref{alg:1})
optimizes over a single column of $A$ (which in effect reduces to fitting a single linear regression
model), and then assigns as the missing values in the corresponding column of $\mbox{M}$ the values
predicted for it by the regression model. While this heuristic for choosing $M$ is quite effective,
it is \textbf{not} a gradient descent step and it therefore leads to a process with unknown
convergence properties. The main motivation for proposing our method was to fix this shortcoming
within the same general framework and propose a method that is similar in spirit, with a convergence
guarantee.

Another advantage of the proposed formulation is the ability to easily extend it to any regularized
linear regression. This can be done by re-writing the itemized form of the objective
(\ref{eq:i_itemized_objective}) as follows:

\[
L(A,M)=\sum_{i}[||(X+M)_{-i}\beta_{i}-(X+M)_{i}||_{F}^{2}+\Omega(\beta_{i})]
\]

\noindent where $\Omega(\beta_{i})$ is the regularization term. 

Now, assuming that the resulting regression problem can be solved (that is, minimizing each of the
summands in the new objective with a constant $M$), and since step (b) of our method remains exactly
the same (the derivative w.r.t $M$ does not change as the extra term does not depend on $M$), we
can use the same method to solve this problem as well.

Another possible extension is to use kernelized linear regression.  This may be useful in cases when
the dependencies between the features are not linear. Here too we can use the same type of method
of optimization, but we defer to future research working out the details of the derivative w.r.t
$M$, which will obviously not remain the same.

The method of initialization is another issue deserving further investigation. Since our procedure
converges to a local minimum of the objective, it may be advantageous to start the procedure from
several random initial points, and choose the best result. However, since the direct target (missing
values) are obviously unknown, we would need an alternative measure of the "goodness" of a
result. Since the missing values are assumed to be missing at random, it would make sense to use the
distance between the distributions of known and imputed values (per feature) as a measure of
appropriateness of an imputation.

\section{\uppercase{Experiments}}
\label{sec:simul}

In order to evaluate our method, we compared its performance to other imputation methods using
various types of data.  We used complete datasets (real or synthetic), and randomly eliminated
entries in order to simulate the missing data case. To evaluate the success of each imputation
method, we used the mean square error (MSE) of the imputed values as a measure of error. MSE is
computed as the mean square distance between stored values (the correct values for the simulated
missing values) and the imputed ones.

In Section~\ref{sec:synthetic} we repeat the experimental evaluation from \cite{templ2011iterative}
using synthetic data, in order to compare the results of our method to the results of IRMI. In
Section~\ref{sec:uci} we compare our method to 3 other methods - IRMI, MI and MICE - using standard
benchmark datasets from the UCI repository \cite{Lichman:2013} . In Section~\ref{sec:storks} we
augment the comparisons with an addition new reallife dataset of storks migration data.

For some real datasets in the experiments described below we report that the IRMI method did not
converge (and therefore did not return any result). This decision was reached when the MSE of the
IRMI method rose at least $6$ orders of magnitude throughout the allocated $50$ iterations, or (when
tested with unlimited iterations) when it rose above the maximum valid number in the system of
approximately $1e+308$.

\subsection {Synthetic data}
\label{sec:synthetic}

\begin{figure*}[!t]
	\centering
	\includegraphics[width=.9\textwidth]{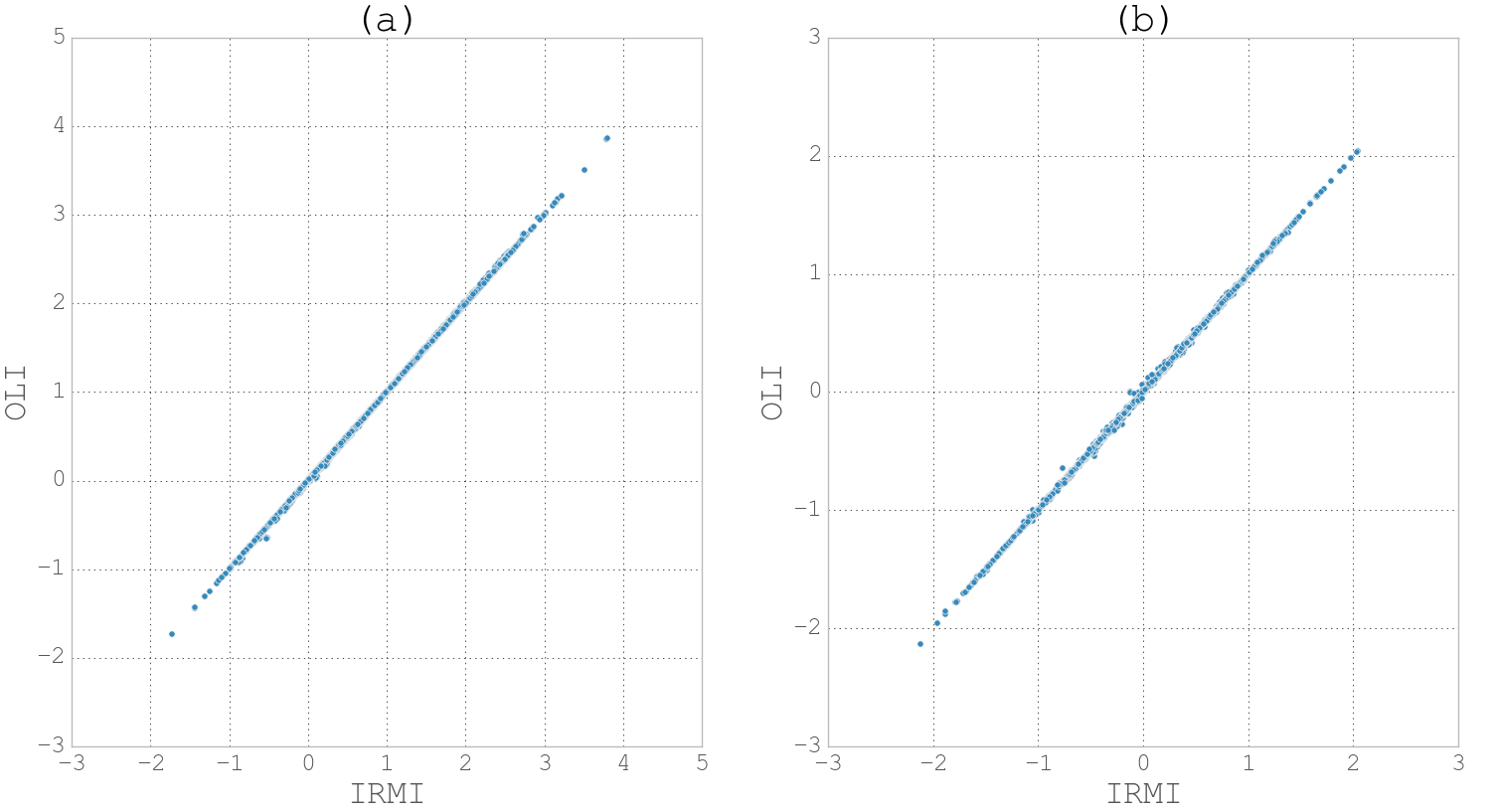}
	
	\protect\caption{(a) Correlation between predicted values for missing data using the IRMI and OLI
		methods. (b) Correlation between the signed error of the prediction for the two methods. }
	\label{fig:1}
\end{figure*}

The following simulation studies follow \cite{templ2011iterative} and compare OLI to IRMI. All
simulations are repeated $20$ times with $10,000$ samples. $5\%$ of all values across records are
selected at random and marked as missing. Values are stored for comparison with imputed
values. Simulation data is multivariate normal with mean of $1$ in all dimensions.  Unless stated
otherwise, the covariance matrix has $1$ in its diagonal entries and $0.7$ in the off-diagonal
entries.

The aim of the first experiment is to test the relationship between the actual values imputed by the
IRMI and OLI methods. The simulation is based on multivariate normal data with $5$
dimensions. Results show that the values imputed by the two methods are highly correlated
(Fig.~\ref{fig:1}a). Furthermore, the signed error ($original-imputed$) is also highly correlated
(Fig.~\ref{fig:1}b). Together, these findings point to the similarity in the results these two
methods produce.

\begin{figure*}[!t]
	\centering
	\includegraphics[width=.92\textwidth, trim={1.2cm 0 0 0}]{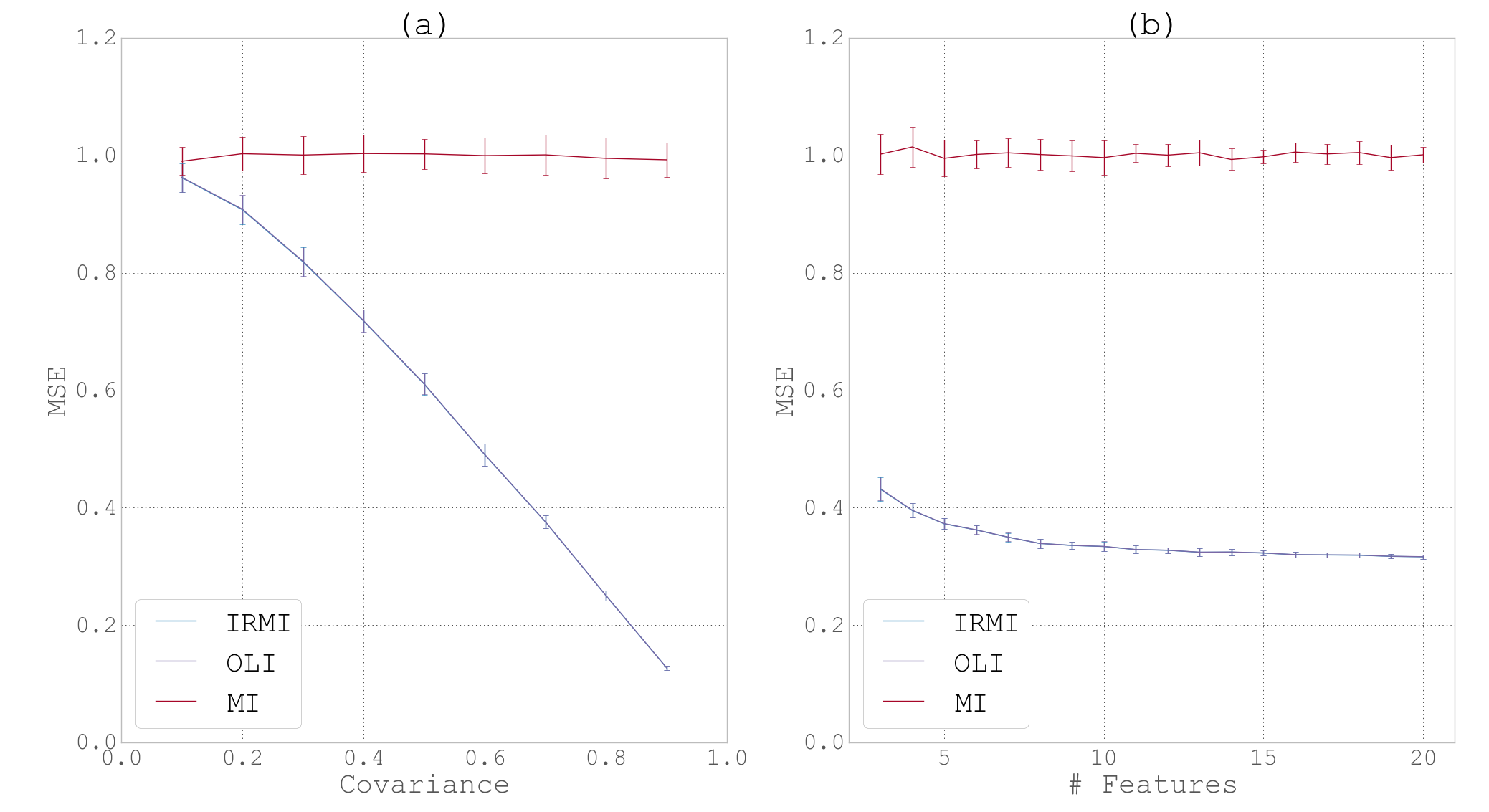}
	
	\protect\caption{(a) MSE of the IRMI, OLI and MI methods as a function of the covariance. Data is
		$5$ dimensional multivariate normal. (b) MSE of the IRMI, OLI and MI methods as a function of the
		dimensionality, with a constant covariance of $0.7$ between pairs of features. In both cases error
		bars represent standard deviation over 20 repetitions. }
	\label{fig:2}
\end{figure*}

\begin{figure}
	\centering
	\includegraphics[width=.5\textwidth, height=.35\textwidth]{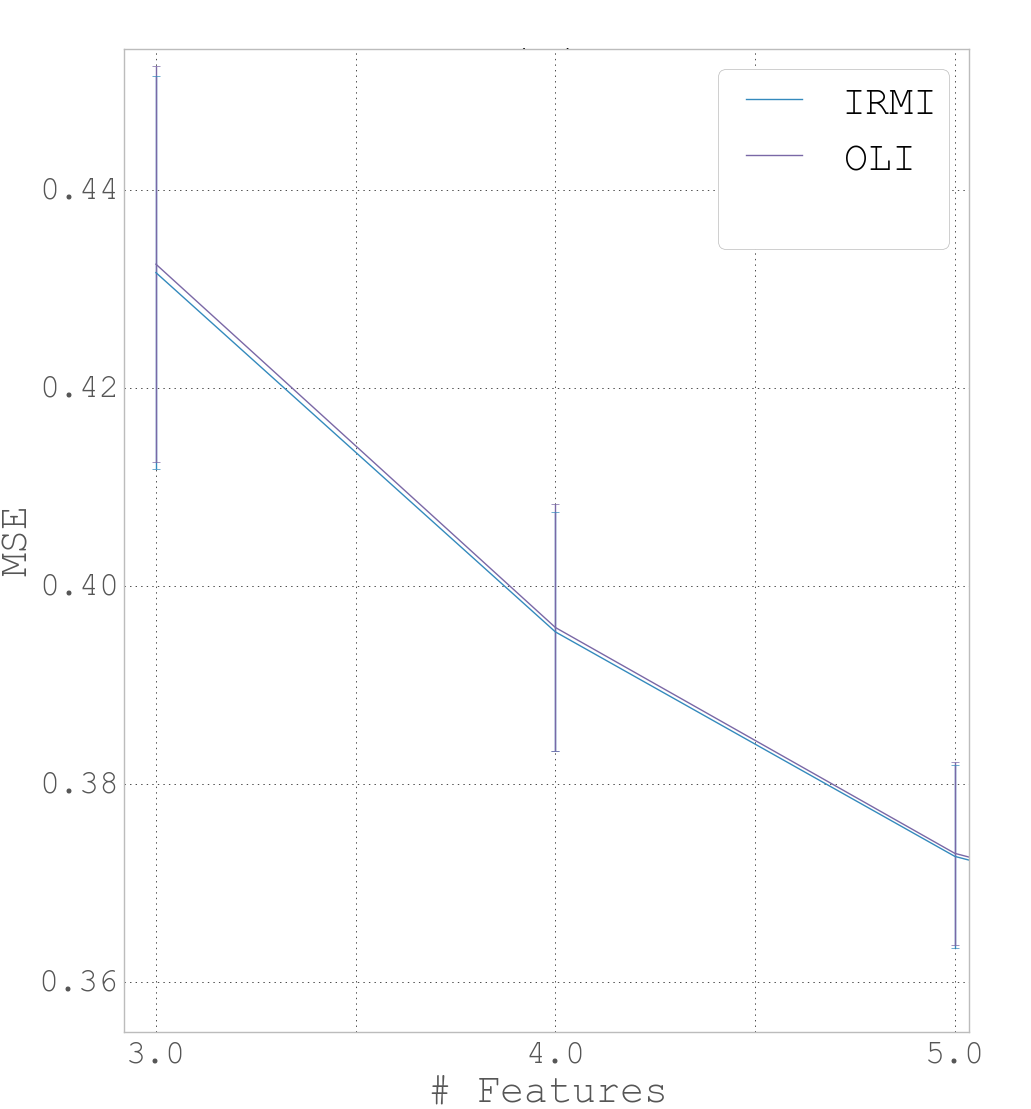}
	\protect\caption{Zoom into a small part of figure \ref{fig:2}}
	\label{fig:3}
\end{figure}

In the next simulation we test the performance of the two methods as we vary the number of
features. The simulation is based on multivariate normal data with $3-20$ dimensions. The results
(Fig.~\ref{fig:2}b) show almost identical behavior of the IRMI and OLI algorithms, which also
coincides with the results presented for IRMI in \cite{templ2011iterative}. Median imputation (MI) is also shown for comparison as baseline. Fig \ref{fig:3} shows a zoom into a small segment of figure \ref{fig:2}. 

As expected, imputing the median (which is also the mean)
of each feature for all missing values results in an MSE equal to the standard deviation of the
features (i.e., $1$). While very close, the IRMI and the OLI methods do not return the exact same
imputation values and errors, with an average absolute deviation of $0.053$

Next we test the performance of the two methods as we vary the covariance between the features. The
simulation is based on multivariate normal data with $5$ dimensions. Non-diagonal elements of the
covariance matrix are set to values in the range $0.1-0.9$. The results (Fig.~\ref{fig:2}a) show
again almost identical behavior of the IRMI and OLI algorithms. As expected, when the dependency
between the feature columns is increased, which is measure by the covariance between the columns
($X$-axis in Fig.~\ref{fig:2}a), the performance of the regression-based methods IRMI and OLI is
monotonically improving, while the performance of the MI method remain unaltered.

\subsection {UCI datasets}
\label{sec:uci}

\begin{table*}[tbh]
	
	\protect\caption{Comparison of the imputation results of the IRMI, OLI, MICE and MI methods with 5\%
		missing data. The \textit{converged} column indicates the number of runs in which the IRMI method
		converged during testing; the MSE of IRMI was calculated for converged repetitions only.}
	\label{tab:1}
	\begin{centering}
		\vspace{0.5cm}
		\begin{tabular}{|l|c|c|c|c|c|c|c|}
			\hline
			Dataset & \# Features & correlation & \multicolumn{2}{c|}{IRMI} & OLI & MI & MICE\tabularnewline
			\cline{4-5}
			&  & & converged & MSE &  &  & \tabularnewline
			\hline
			\hline
			Iris & 4 & 0.59 & 9/10 & \textbf{0.20} & \textbf{0.20} & 1.00 & 0.33\tabularnewline
			\hline
			Ecoli & 7 & 0.18 & 9/10 & 8.26 & 5.75 & 1.72 & \textbf{1.20}\tabularnewline
			\hline
			Wine & 11 & 0.18 & \textbf{0/10} & - & \textbf{0.87} & 1.05 & 1.10\tabularnewline
			\hline
			Housing & 11 & 0.45 & 10/10 & \textbf{0.28} & 0.30 & 1.14 & 0.56\tabularnewline
			\hline
			Power & 4 & 0.45 & \textbf{3/10} & 0.44 & \textbf{0.47} & 1.02 & 0.88 \tabularnewline
			\hline \hline
			Storks & 20 & 0.24 & \textbf{0/10} & - & \textbf{0.31} & 1.07 & 0.42\tabularnewline
			\hline
		\end{tabular} 
		
		\par\end{centering}
	
\end{table*}

The UCI machine learning repository \cite{Lichman:2013} contains several popular benchmark
datasets, some of which have been previously used to compare methods of data imputation
\cite{schmitt2015comparison}.  In the current experiment we used the following datasets:
\textit{iris} \cite{fisher1936use}, \textit{wine} (white) \cite{cortez2009modeling},
\textit{Ecoli} \cite{horton1996probabilistic}, \textit{Boston housing} \cite{harrison1978hedonic},
and \textit{power} \cite{tufekci2014prediction}. Each feature of each dataset was normalized to
have mean $0$ and standard deviation of $1$, in order to make error values comparable between
datasets. Categorical features were dropped. For each dataset, $5\%$ of the values were chosen at
random and replaced with a missing value indicator. The procedure was repeated $10$ times. For these
datasets we also consider the MICE method \cite{buuren2011mice} using the \textit{winMice}
\cite{jacobusse2005winmice} software.

The results are quite good, demonstrating the superior ability of the linear methods to impute
missing data in these datasets (Table~\ref{tab:1}, rows 1-5). In the Iris dataset our OLI method
achieved an average error identical to IRMI, which successfully converged only $9$ out of the $10$
runs. Both outperformed the MI and MICE standard methods. In the Ecoli dataset both the IRMI and OLI
methods performed worse than the alternative methods, with MICE achieving the lowest MSE. In the
Wine dataset the IRMI failed to converge in all $10$ repetitions, while the OLI method outperformed
the MI and MICE methods. The IRMI method outperformed all other methods in the Housing dataset, but
failed to converge $7$ out of $10$ times for the Power dataset.

In summary, in cases where the linear methods were appropriate, with sufficient correlation between
the different features (shown in the second column of Table~\ref{tab:1}), the proposed OLI method
was comparable to the IRMI method with regard to mean square error of the imputed values when the
latter converged, and superior in that it always converges and therefore always returns a result.
While the IRMI method achieved slightly better results than OLI in some cases, its failure to
converge in others gives the OLI method the edge. Overall, better results were achieved for datasets
with high mean correlation between features, as expected when using methods utilizing the linear
relationships between features.

\subsection{Storks behavioral modes dataset}
\label{sec:storks}

In the field of Movement Ecology, readings from accelerometers placed on migrating birds are used for both supervised \cite{resheff2014accelerater} and unsupervised \cite{resheff2015matrix}\cite{resheff2016topic} learning of behavioral modes. In the following experiment we used a dataset of features extracted from $3815$ such measurements. As with the UCI datasets, $10$ repetitions were performed, each with $5\%$ of the values randomly selected and marked as missing. Results (Table 1, final row) of this experiment highlight the relative advantage of the OLI method. While the IRMI method failed to converge in all $10$ repetitions, OLI achieved an average MSE considerably lower than the MI baseline, and also outperformed the MICE method.

\section{\uppercase{Conclusion}}

Since the problem of missing values often haunts real-word datasets while most data analysis methods
are not designed to deal with this problem, imputation is a necessary pre-processing step whenever
discarding entire records is not a viable option. Here we proposed an optimization-based linear
imputation method that augments the IRMI \cite{templ2011iterative} method with the property of
guaranteed convergence, while staying close in spirit to the original method.  Since our method
converges to a local optimum of a different objective function, the two methods should not be
expected to converge to the same value exactly. However, simulation results show that the results of
the proposed method are generally similar (nearly identical) to IRMI when the latter does indeed
converge.

The contribution of our paper is twofold. First, we suggest an optimization problem based method for linear imputation and an algorithm that is guaranteed to converge. Second, we show how this method can be extended to use any number of methods of regularized linear regression. Unlike matrix completion methods \cite{Awagner2015low__matrix_comp}, we do not have a low rank assumption. Thus,  OLI should be preferred when data is expected to have some linear relationships between features and when IRMI fails to converge, or alternatively, when a guarantee of convergence is important (for instance in automated processes). We leave to future research the kernel extension of the OLI method.

\vfill
\bibliographystyle{apalike}
{\small
\bibliography{lib}}

\vfill
\end{document}